\documentclass{article}

\usepackage{booktabs}
\usepackage{colortbl}
\usepackage{amssymb}
\usepackage{times}
\usepackage{epsfig}
\usepackage{graphicx}
\usepackage{amsmath}
\usepackage{amssymb}
\usepackage{multirow}
\usepackage{threeparttable}
\usepackage{caption}

\usepackage{xcolor}

\usepackage[preprint]{corl_2023} 

\title{UniWorld: Autonomous Driving Pre-training via World Models}

\author{
	Chen Min\\
	Peking University\\
	\texttt{minchen@stu.pku.edu.cn} \\
}

\begin{document}
 	\maketitle
	
	
	\begin{abstract}
		In this paper, we draw inspiration from Alberto Elfes' pioneering work in 1989, where he introduced the concept of the occupancy grid as World Models for robots~\cite{occupancy}. We imbue the robot with a spatial-temporal world model, termed UniWorld, to perceive its
		surroundings and predict the future behavior of other participants. 
		UniWorld involves initially predicting 4D geometric occupancy as the World Models for foundational stage and subsequently fine-tuning on downstream tasks. UniWorld can estimate missing information concerning the world state and predict plausible future states of the world. 
		Besides, UniWorld's pre-training process is label-free, enabling the utilization of massive amounts of image-LiDAR pairs to build a Foundational Model.
		The proposed unified pre-training framework demonstrates promising results in key tasks such as motion prediction, multi-camera 3D object detection, and surrounding semantic scene completion. When compared to monocular pre-training methods on the nuScenes dataset, UniWorld shows a significant improvement of about 1.5\% in IoU for motion prediction, 2.0\% in mAP and 2.0\% in NDS for multi-camera 3D object detection, as well as a 3\% increase in mIoU for surrounding semantic scene completion. By adopting our unified pre-training method, a 25\% reduction in 3D training annotation costs can be achieved, offering significant practical value for the implementation of real-world autonomous driving.
		Codes are publicly available at \url{https://github.com/chaytonmin/UniWorld}.
	\end{abstract}

	\section{Introduction}
	\label{sec:intro}

	The multi-camera 3D perception systems in autonomous driving offer a cost-effective solution to gather $360^\circ$ environmental information around vehicles, making it a hot research area recently~\cite{survey1,survey2,bevfusion,chen2023futr3d,li2022unifying}. However, current multi-camera 3D perception models~\cite{detr3d,bevformer,bevdet,bevdepth,m2bev,bevformerv2} usually rely on pre-trained ImageNet models~\cite{imagenet} or depth estimation models~\cite{detr3d} on monocular images. These models fail to take into account the inherent spatial and temporal correlations presented in multi-camera systems. Additionally, while monocular pre-training enhances the capability of image feature extraction, it does not address the pre-training requirements of subsequent tasks. Autonomous driving vehicles collect vast amounts of image-LiDAR pairs, which contain valuable spatial and temporal information. Thus, effectively utilizing these unlabeled image-LiDAR pairs can be beneficial for enhancing the pre-training performance of autonomous driving systems.
	
	\begin{figure*}[t]
		\centering
		\includegraphics[width=0.9\textwidth]{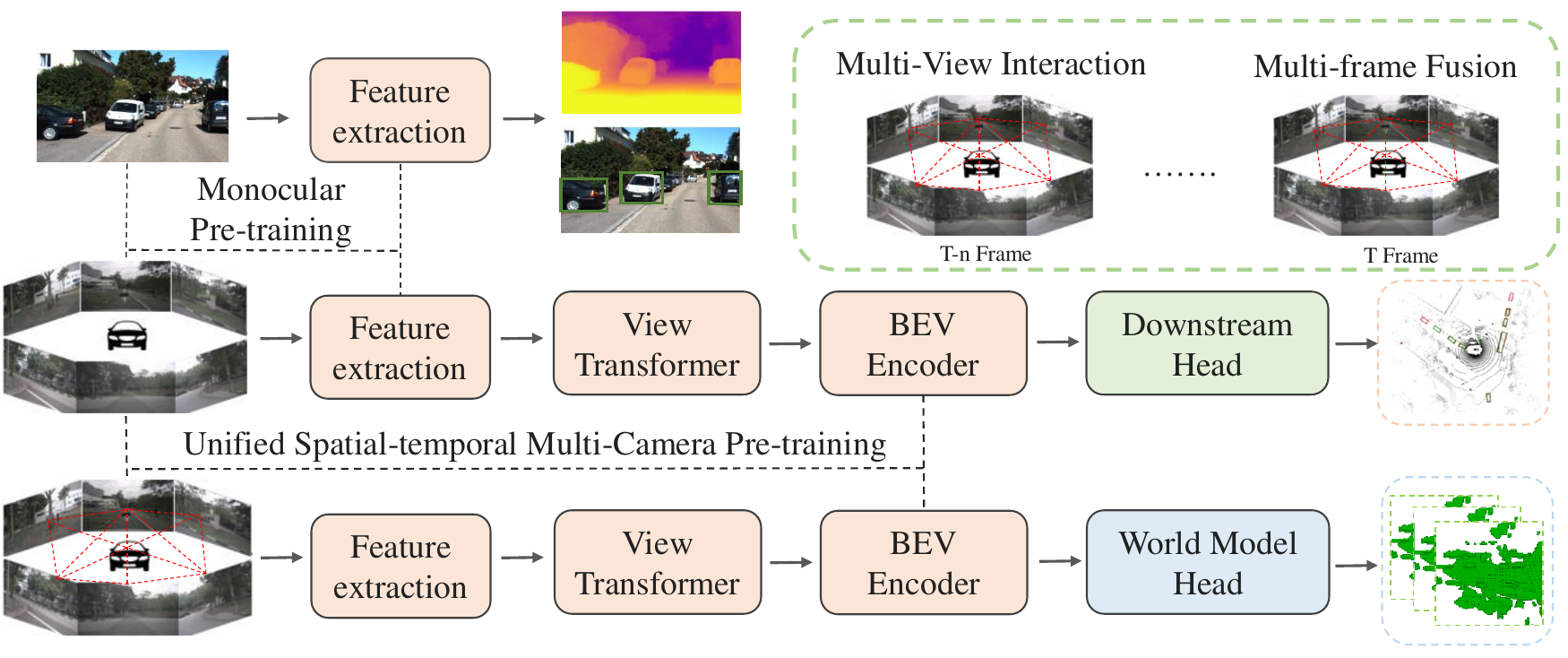} 
		\caption{Comparison between monocular pre-training and our unified multi-camera pre-training. Monocular pre-training only enhances the capability of the feature extraction from a single view, whereas our proposed multi-view unified pre-training enables the incorporation of temporal and spatial information from multi-view images through World Models for pre-training.}
		\label{fig:compare1}
	\end{figure*}
	
	Recent studies, such as BEVDepth~\cite{bevdepth} and DD3D~\cite{dd3d}, have underscored the significance of depth estimation in visual-based perception algorithms. Monocular depth estimation plays a crucial role in acquiring spatial position information for objects. However, depth estimation methods typically focus on estimating the depth of object surfaces, neglecting the holistic 3D structure of objects and occluded elements. For robot systems, geometric occupancy gird provides a unified and consistent world model for robotic tasks, such as obstacle avoidance, path planning, and navigation~\cite{occupancy}. Achieving precise geometric occupancy prediction is instrumental in enhancing the overall 4D perception accuracy within multi-camera perception systems~\cite{occ_survey}. Hence, in the field of autonomous driving, as illustrated in Figure~\ref{fig:compare1}, the pre-training of models would yield greater benefits by prioritizing the reconstruction of the 4D geometric occupancy of the surrounding scene, rather than solely emphasizing depth prediction. 
	
	With an inherent World Model, humans possess the remarkable ability to mentally reconstruct the complete 3D geometry of occluded scenes and anticipate the future motion trajectories of objects in the surrounding scene, which is crucial for recognition and understanding. To imbue the perception system of autonomous vehicles with similar spatial-temporal world models, we propose a multi-camera unified pre-training method, called UniWorld. Our approach leverages the intuitive concept of using the multi-camera system to learn a compressed spatial and temporal representation
	of the environment (i.e., World models) as the foundational stage, followed by fine-tuning downstream tasks. In the case of multi-camera BEV perception, the input multi-camera images are transformed to the BEV space using advanced techniques like LSS~\cite{lss} or Transformer~\cite{detr3d}, and then a geometric occupancy prediction head is incorporated to learn the 4D occupancy distribution, thereby enhancing the model's understanding of the 4D surrounding scene. Due to the sparsity of single-frame point clouds, we employed multi-frame point cloud fusion as the ground truth for 4D occupancy label generation. The decoder was solely used for pre-training, while the well-trained model was utilized to initialize the multi-camera perception models. By designing an effective multi-camera unified pre-training method, we enable the pre-trained model to exploit the rich spatial and temporal information inherent in the unlabeled data. This not only improves the model's ability to understand complex 4D scenes but also reduces reliance on costly and time-consuming manual 3D annotation.
	
	
	To evaluate the effectiveness of our approach, we conducted extensive experiments using the widely used autonomous driving dataset nuScenes~\cite{nuscenes}. The experimental results demonstrate the superiority of our multi-camera unified pre-trained model compared to existing monocular pre-training methods across various 3D perception tasks, including motion prediction, 3D object detection and semantic scene completion.
	For the motion prediction task, our multi-camera unified pre-training algorithm exhibits a 1.8\% increase in IoU and a 1.7\% improvement in VPQ compared to monocular approaches. This indicates that our algorithm is capable of learning future information by constructing World models.
	In the 3D object detection task, the proposed UniWorld-3D achieves a significant improvement of 2.0\% in mAP and 2.0\% in NDS when compared to monocular pre-training methods. This indicates that our model is better equipped to accurately detect and localize objects in a 3D environment.
	For the semantic scene completion task, UniWorld-3D demonstrates a noteworthy improvement of approximately 3\% in mIoU, indicating that our model is more effective in reconstructing and predicting the semantic labels of the surrounding environment. Besides, through the implementation of our integrated pre-training approach, a noteworthy 25\% reduction in costs related to 3D training annotations can be realized. This achievement holds considerable practical significance for the seamless integration of autonomous driving technology into real-world scenarios.	
	The superior performance of our model can be attributed to its ability to effectively leverage unlabeled data, as well as its consideration of spatial and temporal correlations. By incorporating information from multiple camera views, our model can better capture the rich contextual and temporal information present in the scene, leading to enhanced perception capabilities in autonomous driving scenarios. 
	
	The main contributions of this work are listed below:
	\begin{itemize}
		\item We propose to learn spatial-temporal World Models for unified autonomous driving pre-training, which involves initially reconstructing the 4D surrounding scene as the foundamental stage and subsequently fine-tunes on downstream tasks.
		\item UniWorld has the capability to estimate missing information concerning the 3D world state and predict plausible future states of the 4D world. 
		\item UniWorld's pre-training process is label-free, enabling the utilization of massive amounts of image-LiDAR pairs collected by autonomous vehicles to build a Fundational Model.
		\item By adopting our unified pre-training method, a 25\% reduction in costly 3D annotation costs can be achieved, offering significant practical value for the implementation of real-world autonomous driving.
	\end{itemize}
	
	\begin{figure*}[t]
		\centering
 		\includegraphics[width=1\textwidth]{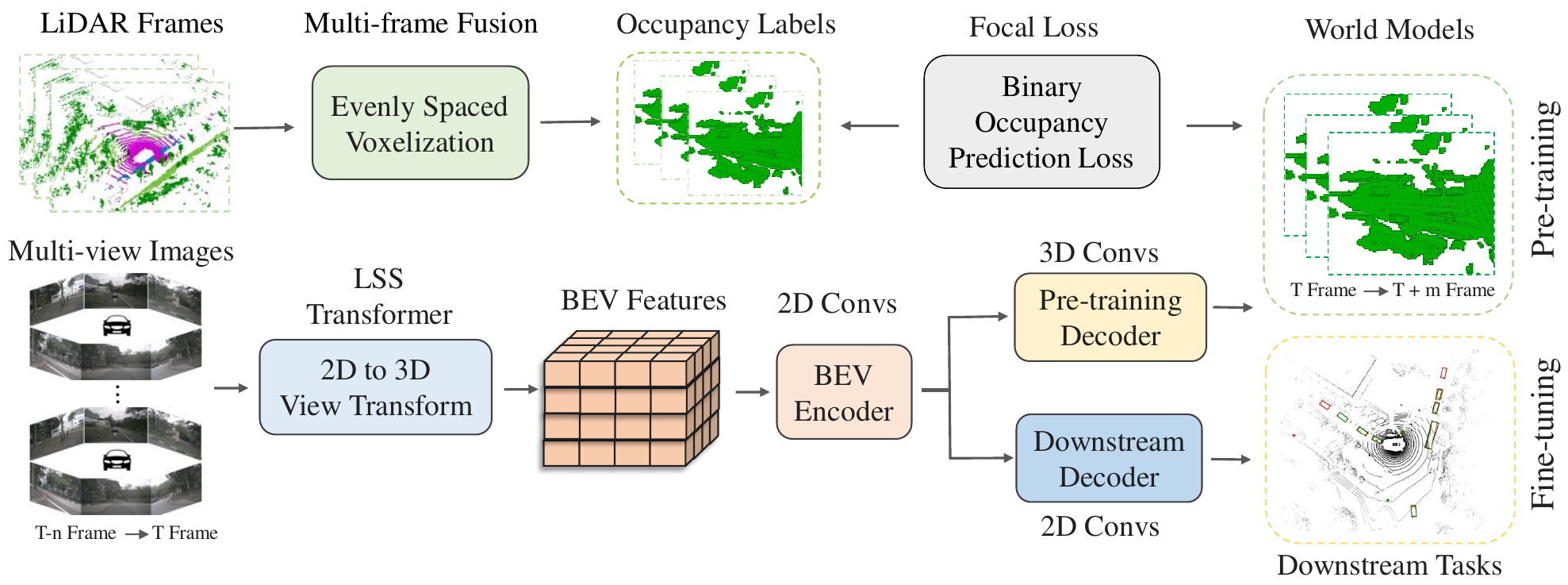}
		\caption{The overall architecture of the proposed multi-camera unified pre-training method UniWorld. We first transform the multi-frame large-scale irregular LiDAR point clouds into volumetric representations as the 4D geometric occupancy labels, then add an occupancy decoder with some layers of 3D convolutions to the BEV encoder. We apply binary occupancy classification as the pretext task to distinguish whether the 4D voxel contains points. After pre-training, the lightweight decoder is discarded, and the encoder is used to warm up the backbones of downstream tasks.
		}
		\label{fig:flowchart}
	\end{figure*}
	
	\section{Related Work}
	\subsection{Multi-Camera 3D Perception}
	
	In the field of autonomous driving, vision-based 3D perception conducted in bird's eye view has gained significant attention in recent years~\cite{zhang2022beverse,jiang2022polarformer,hu2022st,xie2023robobev,chen2022graph}. Learning-based BEV perception methods, based on 2D-to-3D view transformation, can be broadly categorized into geometry-based and Transformer-based approaches~\cite{fastbev,li2022hdmapnet,liao2022maptr,doll2022spatialdetr,zhou2022matrixvt}. One of the early geometry-based methods is LSS \cite{lss}, which lifts each individual image into a frustum of features for each camera and then combines them into a rasterized BEV grid. Building upon LSS, BEVDet \cite{bevdet} introduces image-view and BEV space data augmentation techniques. 
	BEVDepth \cite{bevdepth} demonstrates the significance of depth and improves the quality of BEV features by incorporating explicit depth supervision from LiDAR. BEVStereo \cite{bevstereo} and STS \cite{sts} leverage temporal multi-view stereo methods to enhance depth precision. SOLOFusion~\cite{solofusion} and VideoBEV~\cite{videobev} explore long-term temporal fusion for multi-view 3D perception. DETR3D \cite{detr3d} is the first Transformer-based BEV method, which defines object queries in 3D space and learns from multi-view image features using a transformer decoder. Building upon DETR3D, PETR \cite{petr} enhances the approach with position embedding transformations, while BEVFormer \cite{bevformer} introduces temporal self-attention to fuse historical BEV features. UniAD \cite{uniad} extends BEVFormer to enable multi-task learning in BEV space. Although the existing BEV perception methods have shown promising performance, they are typically initialized with ImageNet pre-trained models \cite{imagenet} or depth pre-trained models \cite{dd3d} trained on monocular images. However, there is a lack of unified pre-training methods that effectively leverage the geometric structure of multi-camera inputs.
	
	\subsection{Label-free Pre-training}
	
	Label-free Pre-training has gained significant popularity in recent years as it eliminates the need for expensive data annotation. For instance, the method presented in \cite{patch_id} focuses on predicting the relative location of image patches as the pretext task. Another approach, as described in \cite{jigsaw}, tackles a jigsaw puzzle prediction task, which demonstrates strong generalization capabilities for domain adaptation in object recognition. DeepCluster \cite{deepcluster} and SwAV \cite{swav} leverage k-means clustering to obtain pseudo-labels, which are then used to train the network. Moco \cite{moco} and BYOL \cite{byol} construct contrastive views for self-supervised learning. Additionally, methods like MAE \cite{mae} and BEiT \cite{beit} employ a random patch masking approach where missing pixels or features are reconstructed using a simple autoencoder framework. In the context of automated vehicle perception, DD3D utilizes monocular depth estimation for pre-training. Voxel-MAE \cite{voxel-mae} and ALSO~\cite{also} propose predicting occupancy for LiDAR perception as the pretext task. Our previous work Occ-BEV~\cite{occbev} defines the task of multi-camera unified pre-training and reconstructes the 3D static surrounding scene as
	the fundamental stage. In this work, we extend Occ-BEV to utilize 4D geometric occupancy prediction to learn spatial-temporal world models for vision-based perception.
	
	\subsection{World Models}
	The concept of employing world models by humans, animals, and intelligent systems has a historical foundation in psychology, dating back to the work of Craik in 1943~\cite{achinstein1983nature}. Alberto Elfes proposed the geometric occupancy grid as a world model for robot perception and navigation in 1989~\cite{occupancy}. David Ha proposed that world model can be trained
	quickly in an unsupervised manner to learn a
	compressed spatial and temporal representation
	of the environment~\cite{world_models}. Methods in ~\cite{recurrent_wm,mastering_wm,mastering_wm2} presuppose access to rewards and online interaction with the environment from predictions in the
	compact latent space of a world model in reinforcement learning. ~\cite{babaeizadeh2017stochastic,denton2018stochastic,franceschi2020stochastic} learned the latent dynamics of a world model from image observations via
	video prediction. MILE~\cite{hu2022model} proposed to build the world model by predicting the future BEV
	segmentation from high-resolution videos of
	expert demonstrations for autonomous driving. In this paper, we follow the concept of the occupancy grid as a robot's world model introduced in 1989~\cite{occupancy}, and propose a label-free spatiotemporal fused world model for autonomous driving by integrating future prediction techniques inspired by methods in reinforcement learning~\cite{world_models} and MILE~\cite{hu2022model}.
	\section{Methodology}
	
	This section provides a detailed description of the network architecture employed by UniWorld, depicted in Figure~\ref{fig:flowchart}. We commence by examining the vision-based bird's eye view (BEV) perception methods in Section~\ref{review}. Following that, in Section~\ref{geo}, we present the proposed 4D world model utilized for pre-training. Additionally, we compare this model to existing monocular pre-training, knowledge distillation, and 3D static pre-training methods in Section~\ref{com}.
	
	\subsection{Review of BEV Perception}
	\label{review}
	
	As discussed in the related works, there are two primary learning-based approaches for transforming 2D images into 3D space: LSS-based~\cite{lss} and Transformer-based~\cite{detr3d} view transformations. However, our method is not restricted to any specific view transformation technique. In the subsequent sections, we will provide a comprehensive outline of the workflow for multi-camera perception algorithms based on the bird's eye view.
	
	The input images from multiple cameras, denoted as $I=\{I_i, i=1, 2, ..., N_{view}\}$, are initially processed by an image backbone network, such as ResNet-101~\cite{resnet}, which generates feature maps $F_{2d}=\{F_{2d}^{i}\}_{i=1}^{N_{view}}$ for each camera view. These features are then passed through a 2D-to-3D view transformation operation, projecting them onto a unified bird's eye view representation denoted as $F_{bev}\in \mathbb{R}^{C\times H\times W}$. By incorporating specific heads, various autonomous driving perception tasks can be performed on the bird's eye view, including 3D object detection, map segmentation, object tracking, and more~\cite{uniad}.
	
	Current bird's eye view perception algorithms~\cite{bevformer,detr3d,bevdet} often rely on feature extraction models (e.g., ImageNet~\cite{imagenet}) or depth estimation models (e.g., V2-99~\cite{dd3d}) trained on monocular images. However, these approaches do not consider the interplay and correlation between images captured from different camera views and frames, leading to a lack of a unified pre-training model that utilizes the spatial and temporal relationships between different camera views. In order to fully exploit these relationships, we propose a multi-camera unified pre-training model.
	
	
	Methods such as BEVDepth~\cite{bevdepth} and DD3D~\cite{dd3d} demonstrate the importance of depth estimation for visual-based perception algorithms. However, depth estimation can only estimate the position of the object's surface, ignoring the occlusions of objects. For multi-camera systems, precision occupancy prediction is beneficial to the accuracy of perception. To enable the model to possess the capabilities of occupancy scene completion and future prediction simultaneously, we propose to build world models for multi-camera unified pre-training via 4D geometric occupancy prediction.
	
	\subsubsection{4D Geometric Occupancy Decoder}	
	In order to predict 4D geometric occupancy using the BEV features $F_{bev}$, we begin by transforming the BEV features into $F_{bev}^{'}\in \mathbb{R}^{C^{'}\times D\times H\times W}$, where D represents the number of height channels, and $C=C^{'}\times D$. Subsequently, we utilize a 3D decoder specifically designed to generate 4D geometric occupancy. This decoder consists of lightweight 3D convolution layers, with the final layer providing the probability of each voxel containing points. The output of the decoder is denoted as $P=\{P^{i}\in \mathbb{R}^{D\times H \times W \times 1}, i=1, 2, ..., m \}$. During the pre-training phase, the main objective of the decoder is to reconstruct occupied voxels.
	
	\subsubsection{Pre-training Target}
	\label{geo}
	Considering the sparsity of single-frame LiDAR point clouds and the potential inaccuracies that arise from fusing a large number of frames due to dynamic objects, we adopt a strategy of fusing LiDAR point clouds from selected keyframes to generate occupancy labels. Following the standard practice in 3D perception models~\cite{second,pv_rcnn,centerpoint,cylinder3d}, the LiDAR point clouds are divided into evenly spaced voxels. For the dimensions of the LiDAR point clouds along $Z\times Y\times X$ (denoted as $D\times H\times W$), the voxel size is determined as $v_Z\times v_H\times v_W$ respectively. The 4D ground truth $T=\{T^{i} \in \{0,1\}^{D \times H\times W\times 1}, i=1, 2, ..., m\}$ is generated based on the occupancy of the voxels, indicating whether each voxel contains points or not. A value of 1 represents an occupied voxel, while a value of 0 indicates a free voxel.
	
	We propose the binary geometric occupancy classification task as part of the pre-training process for multi-camera perception models. The goal of this task is to train the network to accurately predict the distribution of geometric occupancy in a 4D scene based on multi-view images. However, due to the significant number of empty voxels, predicting occupancy grids presents an imbalanced binary classification challenge. To address this, we employ focal loss for binary occupancy classification, leveraging the predicted 4D occupied values $P$ and the 4D ground truth occupied voxels $T$:
	
	\begin{equation} \label{loss}
	loss = -\frac{1}{m}\frac{1}{n}\sum_{i=1}^{m}\sum_{j=1}^{n}\alpha_t \left(1 - P_t^{ij}\right)^\gamma \log(P_t^{ij}),
	\end{equation}
	where $P^{ij}$ is the predicted probability of voxel $j$ in the $i$-th
	occupancy. $n=D\times H \times W$ is the total number of voxels and batch is the number of batch sizes. The weighting factor $\alpha$ for positive/negative examples is set as 2 and the weighting factor $\gamma$ for easy/hard examples is 0.25. $\alpha_t=\alpha$ and $P_t^{ij} = P^{ij}$ for class 1. $\alpha_t=1-\alpha$ and $P_t^{ij} = 1-P^{ij}$ for class 0. 
	
	Currently, the 4D geometry occupancy labels used in the algorithm are obtained from multi-frame LiDAR point clouds. In the future, it is also feasible to utilize point clouds generated from 3D scene reconstructions using techniques such as NeRF~\cite{nerf,nerf2,block-nerf,dnerf} or MVS~\cite{mvs,mvsnet,aa-rmvsnet,bi}.

	\subsubsection{Pre-training for Surrounding Semantic Occupancy Prediction}
	
	Recently, several algorithms, such as TPVFormer~\cite{tpvformer}, OpenOccupancy~\cite{openoccupancy}, and Occ3D~\cite{occ3d}, have expanded the scope of multi-camera BEV perception to include the task of surrounding semantic scene completion~\cite{li2023voxformer,wei2023surroundocc,zhang2023occformer,miao2023occdepth,gan2023simple}. However, directly predicting the 3D semantics of multi-view images requires a significant amount of 3D semantic annotations for training, which can be expensive and time-consuming. To overcome this challenge, we propose to extend our multi-camera unified pre-training algorithm to include the surrounding semantic scene completion task. This involves initially performing geometric occupancy prediction and subsequently fine-tuning the model on the semantic scene completion task.
	
	
	\subsection{Comparision with Existing Methods}
	\label{com}
	\subsubsection{Comparision with Monocular Pre-training}
	
	Currently, multi-camera perception algorithms predominantly utilize either monocular image pre-training on ImageNet~\cite{imagenet} or depth estimation pre-training~\cite{dd3d}. In Figure~\ref{fig:compare1}, we highlight several advantages of our proposed multi-camera unified pre-training model over monocular pre-training:
	(1) \textbf{Spatial-Temporal Integration}: By leveraging spatial and temporal information from multiple camera views, the model gains a better understanding of the dynamic nature of the environment, leading to more accurate predictions.
	(2) \textbf{Unified Representation}: The unified pre-training approach enables the model to learn a shared representation across different camera views, facilitating improved knowledge transfer and reducing the necessity for task-specific pre-training.
	(3) \textbf{Perception of Occluded Areas}: Monocular depth estimation can only predict surface positions of objects, while our proposed multi-camera unified pre-training method enables comprehensive 3D reconstruction of occluded objects.
	
%
%
	\subsubsection{Comparision with Knowledge Distillation}
	Recently, there have been advancements in knowledge distillation algorithms such as BEVDistill~\cite{bevdistill}, TiG-BEV~\cite{tigbev} and GeoMIM~\cite{geomim}, which aim to transfer knowledge from well-established 3D LiDAR models like CenterPoint~\cite{centerpoint} to multi-camera object detection algorithms. Similarly, our approach aims to leverage the rich spatial information presented in 3D point clouds and transfer it to multi-camera algorithms. Our unique pre-training algorithm eliminates the need for annotations or pre-trained LiDAR detection models, significantly reducing the 3D annotation requirements.
	
	\section{Experiments}
	
		\begin{table*}[t]
		\centering
		\renewcommand{\arraystretch}{1.5}
		\caption{Quantitative multi-task learning performance on the nuScenes validation set.}
		\resizebox{\textwidth}{!}
		{
			{
				\begin{tabular}{c|c|cc|cccc|cc}
					\toprule
					\multirow{2}*{\textbf{Method}}&\multirow{2}*{\textbf{Pre-train}}&\multicolumn{2}{c|}{\textbf{Detection}}&\multicolumn{4}{c|}{\textbf{Semantic map}}&\multicolumn{2}{c}{\textbf{Motion}} \\
					&&\textbf{NDS}$\uparrow$ & \textbf{MAP}$\uparrow$ &\textbf{Divider}$\uparrow$&\textbf{Ped Cross}$\uparrow$&\textbf{Boundary}$\uparrow$&\textbf{mIoU}$\uparrow$ &\textbf{IoU}$\uparrow$&\textbf{VPQ}$\uparrow$ \\
					\midrule
					\multirow{3}*{ BEVerse~\cite{beverse}}&ImageNet~\cite{imagenet}&0.466& 0.321& 53.2 &39.0& 53.9 &48.7& 38.7& 33.3 \\
					\cline{2-10}
					&ImageNet + UniWorld-3D&$0.483^{\textcolor{teal} {+1.7\%}}$& $\textbf{0.334}^{\textcolor{teal} {+1.3\%}}$&$ \textbf{54.8} ^{\textcolor{teal} {+1.6}}$&$\textbf{40.4} ^{\textcolor{teal} {+1.4}}$&$ 55.0 ^{\textcolor{teal} {+1.1}}$&$\textbf{50.0}^{\textcolor{teal} {+1.3}}$&$ 39.4 ^{\textcolor{teal} {+0.7}}$&$ 34.1 ^{\textcolor{teal} {+0.8}}$\\
					\cline{2-10}
					&ImageNet + UniWorld-4D&$\textbf{0.484}^{\textcolor{teal} {+1.8\%}}$&$ 0.331^{\textcolor{teal} {+1.0 \%}}$&$ 54.7 ^{\textcolor{teal} {+1.5}}$&$40.1^{\textcolor{teal} {+1.1}}$&$ 54.9 ^{\textcolor{teal} {+1.0}}$&$49.9^{\textcolor{teal} {+1.2}}$&$ \textbf{40.5}^{\textcolor{teal} {+1.8}}$&$ \textbf{35.0}^{\textcolor{teal} {+1.7}}$\\
					\bottomrule
				\end{tabular}
		}}
		\label{tab:beverse}
	\end{table*}
	
	\begin{table*}[t]
		\centering
		\renewcommand{\arraystretch}{1.5} 
		\caption{Quantitative multi-camera 3D object detection performance on the nuScenes validation set.}
		\resizebox{\textwidth}{!}
		{
			{
				\begin{tabular}{c|c|c|c|c|c|c|c|c|c|c|c|c}
					\toprule
					\textbf{Type}&\textbf{Method}&\textbf{Pre-train}&\textbf{Backbone}&\textbf{Image Size}&\textbf{CBGS} &\textbf{mAP}$\uparrow$&\textbf{NDS}$\uparrow$&\textbf{mATE}$\downarrow$&\textbf{mASE}$\downarrow$&\textbf{mAOE}$\downarrow$&\textbf{mAVE}$\downarrow$&\textbf{mAAE}$\downarrow$ \\
					\midrule
					\multirow{8}*{Transformer~\cite{detr3d}}	
					&\multirow{2}*{DETR3D~\cite{detr3d}}&FCOS3D~\cite{fcos3d}&R101-DCN&900$\times$1600&$\checkmark$ &0.349&0.434&0.716&0.268&0.379&0.842&0.200\\
					\cline{3-13}
					&&FCOS3D + UniWorld-3D &R101-DCN&900$\times$1600&$\checkmark$ &$\textbf{0.360}^{\textcolor{teal} {+1.1\%}}$&$\textbf{0.461}^{\textcolor{teal} {+2.7\%}}$&0.701&0.260&0.372&0.730&0.188\\
					
					\cline{2-13}
					
					&\multirow{6}*{BEVFormer~\cite{bevformer}}&ImageNet~\cite{imagenet}&R101-DCN&900$\times$1600&$\times$ &0.371&0.479&0.688&0.275&0.441&0.382&0.195\\
					\cline{3-13}
					&&ImageNe+UniWorld-3D&R101-DCN&900$\times$1600&$\times$ &$\textbf{0.397}^{\textcolor{teal} {+2.6\%}}$&$0.500^{\textcolor{teal} {+2.1\%}}$&0.680&0.272&0.392&0.380&0.193\\
					\cline{3-13}
					&&ImageNe+UniWorld-4D&R101-DCN&900$\times$1600&$\times$ &$0.390^{\textcolor{teal} {+1.9\%}}$&$\textbf{0.501}^{\textcolor{teal} {+2.2\%}}$&0.683&0.280&0.394&0.374&0.192\\
					\cline{3-13}
					
					&&FCOS3D~\cite{fcos3d}&R101-DCN&900$\times$1600&$\times$ &0.416&0.517&0.673&0.274&0.372&0.394&0.198\\
					\cline{3-13}
					&&FCOS3D + UniWorld-3D&R101-DCN&900$\times$1600&$\times$ &$\textbf{0.438}^{\textcolor{teal} {+2.2\%}}$&$\textbf{0.534}^{\textcolor{teal} {+1.7\%}}$&0.656&0.271&0.371&0.348&0.183\\
					\cline{3-13}
					&&FCOS3D + UniWorld-4D&R101-DCN&900$\times$1600&$\times$ &$0.432^{\textcolor{teal} {+1.6\%}}$&$0.530^{\textcolor{teal} {+1.3 \%}}$&0.659&0.274&0.375&0.344&0.188\\
					
%
					\bottomrule
				\end{tabular}
		}}
		\label{tab:nuscenes_val}
	\end{table*}

	\begin{table*}[t]
		\centering
		\renewcommand{\arraystretch}{1.5}
		\caption{Quantitative multi-camera 3D object detection performance on the nuScenes test set.}
		\resizebox{\textwidth}{!}
		{
			{
				\begin{tabular}{c|c|c|c|c|c|c|c|c|c|c|c}
					\toprule
					\textbf{Method}&\textbf{Pre-train}&\textbf{Backbone}&\textbf{Image Size}&\textbf{CBGS} &\textbf{mAP}$\uparrow$&\textbf{NDS}$\uparrow$&\textbf{mATE}$\downarrow$&\textbf{mASE}$\downarrow$&\textbf{mAOE}$\downarrow$&\textbf{mAVE}$\downarrow$&\textbf{mAAE}$\downarrow$ \\
					\midrule
				 \multirow{2}*{ DETR3D~\cite{detr3d}}&DD3D~\cite{dd3d}&v2-99&900$\times$1600&$\checkmark$&0.412&0.479&0.641 &\textbf{0.255} &\textbf{0.394} &0.845 &0.133\\
					\cline{2-12}
					&DD3D + UniWorld-3D&v2-99&900$\times$1600&$\checkmark$&$\textbf{0.431}^{\textcolor{teal} {+1.9\%}}$&$\textbf{0.496}^{\textcolor{teal} {+1.7\%}}$&\textbf{0.621}&0.257&0.407&\textbf{0.783}&\textbf{0.123}\\
					\bottomrule
				\end{tabular}
		}}
		\label{tab:nuscenes_test}
	\end{table*}

	\begin{table*}[t]
		\centering
		\renewcommand{\arraystretch}{1.5}
		\caption{Quantitative segmentation performance on the 3D occupancy prediction challenge~\cite{occ}.}
		\resizebox{\textwidth}{!}{
			{
				\begin{tabular}{c|c|c|c|c|c|c|c|c|c|c|c|c|c|c|c|c|c|c|c|c|c}
					\toprule
					\textbf{Methods} &\textbf{Pre-train}&\textbf{Backbone}&\textbf{Image Size}&\textbf{mIoU}$\uparrow$ &\textbf{\rotatebox{90}{others}}&\textbf{\rotatebox{90}{barrier}}&\textbf{\rotatebox{90}{bicycle}}&\textbf{\rotatebox{90}{bus}}&\textbf{\rotatebox{90}{car}}&\textbf{\rotatebox{90}{construction}}&\textbf{\rotatebox{90}{motorcycle}}&\textbf{\rotatebox{90}{pedestrian}}&\textbf{\rotatebox{90}{traffic-cone}}&\textbf{\rotatebox{90}{trailer}}&\textbf{\rotatebox{90}{truck}}&\textbf{\rotatebox{90}{driveable}}&\textbf{\rotatebox{90}{others}}&\textbf{\rotatebox{90}{sidewalk}}&\textbf{\rotatebox{90}{terrain}}&\textbf{\rotatebox{90}{manmade}}&\textbf{\rotatebox{90}{vegetation}}\\
					\midrule
					BEVFormer~\cite{bevformer} &FCOS3D~\cite{fcos3d} &R101-DCN~\cite{resnet}&900$\times$1600&23.70&	10.24&	36.77&	11.70&	29.87&	38.92&	10.29&	22.05&	16.21&	14.69&	27.44&	23.13&	48.19&	33.10&	29.80&	17.64&	19.01&	13.75\\
					\midrule
					BEVStereo~\cite{bevstereo}&BEVDet4D~\cite{bevdet} &Swin-B~\cite{swin}&512$\times$1408&42.78&	22.45&	47.95&	28.13&	40.29&	53.79&	27.60&	35.18&	29.64&	31.69&	45.49&	37.71&	81.88&	49.16&	55.03&	51.00&	50.87&	39.44\\
					\midrule
					BEVStereo~\cite{bevstereo}&UniWorld-3D &Swin-B~\cite{swin}&512$\times$1408&$\textbf{45.92}^{\textcolor{teal} {+3.14\%}}$&\textbf{26.21}&	\textbf{53.06}&	\textbf{33.41}&	\textbf{42.77}&	\textbf{56.57}&	\textbf{28.99}&	\textbf{39.92}&	\textbf{32.31}&	\textbf{34.89}&	\textbf{49.59}&	\textbf{40.28}&	\textbf{82.88}&	\textbf{52.29}&	\textbf{57.77}&	\textbf{53.58}&	\textbf{53.94}&	\textbf{42.25}
					\\
					\bottomrule
				\end{tabular}
		}}
		\label{tab:nuscenes_seg}
	\end{table*}
	
	\subsection{Experimental Setup}
	%
	We conducted extensive experiments on the nuScenes dataset~\cite{nuscenes}. 
	We adopted the training settings from the existing methods: DETR3D~\cite{detr3d} and BEVFormer~\cite{bevformer}, which are two Transformer-based methods, and BEVerse~\cite{beverse}, BEVDet~\cite{bevdet}, BEVDepth~\cite{bevdepth}, and BEVStereo~\cite{bevstereo}, which are four LSS-based methods. We performed pre-training for a total of 24 epochs. The occupancy decoder consists of two layers of 3D convolutional layers. For more detailed information about the parameter setups, please refer to the papers of DETR3D, BEVFormer, BEVerse, BEVDet, BEVDepth and BEVStereo. All experiments were conducted using 8 Nvidia Tesla A40 GPU cards.
	\subsection{Results on Downstream Tasks}
	\subsubsection{Multi-Camera 3D Object Detection}
	
	 We first comprehensively validate the performance of the proposed multi-camera unified pretraining algorithm on a multi-task model, i.e., BEVerse~\cite{beverse}. As shown in Table~\ref{tab:beverse}, compared to the baseline, both Occ-BEV~\cite{occbev} and UniWorld significantly enhance the performance of 3D object detection, semantic
	 map construction, and motion prediction. In comparison to Occ-BEV's 3D reconstruction pre-training, UniWorld introduces a 4D occupancy prediction auxiliary task, enabling the model to learn 4D motion information and construct a more comprehensive 4D world model. In the motion prediction task, UniWorld's improvement over Occ-BEV results in a 1\% increase in IoU and VPQ.
	
	We subsequently performed an assessment of the performance of multi-camera unified pre-training models in the 3D object detection task using the validation set of nuScenes. As shown in Table~\ref{tab:nuscenes_val}, our multi-camera unified pre-training methods Occ-BEV and UniWorld exhibited significant improvements over monocular FCOS3D~\cite{fcos3d}. Occ-BEV surpassed FCOS3D~\cite{fcos3d} on DETR3D~\cite{detr3d} by achieving a 2.7\% increase in NDS and 1.1\% in mAP. Additionally, it outperformed BEVFormer~\cite{bevformer} with a 2.6\% improvement in NDS and 2.1\% in mAP for ImageNet baseline, a 1.7\% improvement in NDS and 2.2\% in mAP for FCOS3D baseline. We present the convergence curve of BEVFormer~\cite{bevformer} in Figure~\ref{fig:epoch} for FCOS3D baseline. Our unified pre-training Occ-BEV significantly enhances BEVFormer~\cite{bevformer} at the initial epoch, achieving a 4\% increase in NDS. This demonstrates that our unified pre-training method delivers accurate object position information from a global perspective. 
	
	Compared to the 3D reconstruction pretraining of Occ-BEV~\cite{occbev}, the introduction of 4D occupancy prediction in UniWorld resulted in slightly inferior performance in the 3D object detection task. This discrepancy is likely attributed to the fact that 4D prediction allows the model to learn future information but introduces uncertainty in the object's position. Further resolution of this issue is required.
	
	
	For further validation, we conducted additional experiments on the nuScenes test set to validate the effectiveness of our proposed multi-camera unified pre-training method Occ-BEV via 3D scene reconstruction compared to pre-training based on monocular depth estimation. As presented in Table~\ref{tab:nuscenes_test}, our multi-camera unified pre-training method demonstrated a significant improvement of about 1.8\% in both mAP and NDS compared to the DETR3D~\cite{detr3d} pre-trained on DD3D~\cite{dd3d} for depth estimation. This highlights the effectiveness and superiority of our pre-training approach in enhancing the performance of 3D perception tasks. Compared to the monocular depth estimation approach of DD3D~\cite{dd3d}, our pre-training method considers the complete 3D structure of objects, beyond the partial surfaces captured by LiDAR.
	Moreover, it incorporates the learning of multi-view and temporal information, allowing for a more comprehensive understanding of the scene. 
	The above results indicated that our proposed Occ-BEV model has a promising application in autonomous driving.	As UniWorld's performance in the 3D object detection task is slightly inferior to Occ-BEV, we do not present its results on the test set here.
	
	
	We also compared our proposed multi-camera unified pre-training method Occ-BEV with the knowledge distillation approach BEVDistill~\cite{bevdistill}. As shown in Table~\ref{tab:kd}, our method demonstrates comparable performance to the knowledge distillation method trained on annotated LiDAR point clouds data. It is worth noting that our approach offers higher efficiency and broader applicability since it does not rely on data annotation or training of LiDAR point clouds models as BEVDistill~\cite{bevdistill}.
	
	\subsubsection{Multi-Camera Semantic Occupancy Prediction}
	
	We also evaluated the performance of our proposed multi-camera unified pre-training method Occ-BEV on the task of multi-camera semantic scene completion. Compared to BEV perception, the task of predicting semantic labels for each voxel in 3D space, known as surrounding semantic scene completion, is more challenging. To tackle this challenge, we decomposed the task into two steps: first reconstructing the 3D scene as the fundamental model and then simultaneously reconstructing and predicting semantics. As shown in Table~\ref{tab:nuscenes_seg}, on the 3D occupancy prediction challenge~\cite{occ}, our algorithm achieved a 3\% improvement in mIoU compared to BEVStereo~\cite{bevstereo}, highlighting the effectiveness of our approach in addressing the complexities of surrounding semantic occupancy prediction.
	
	\begin{minipage}{.33\linewidth} 
		\includegraphics[width=.99\linewidth]{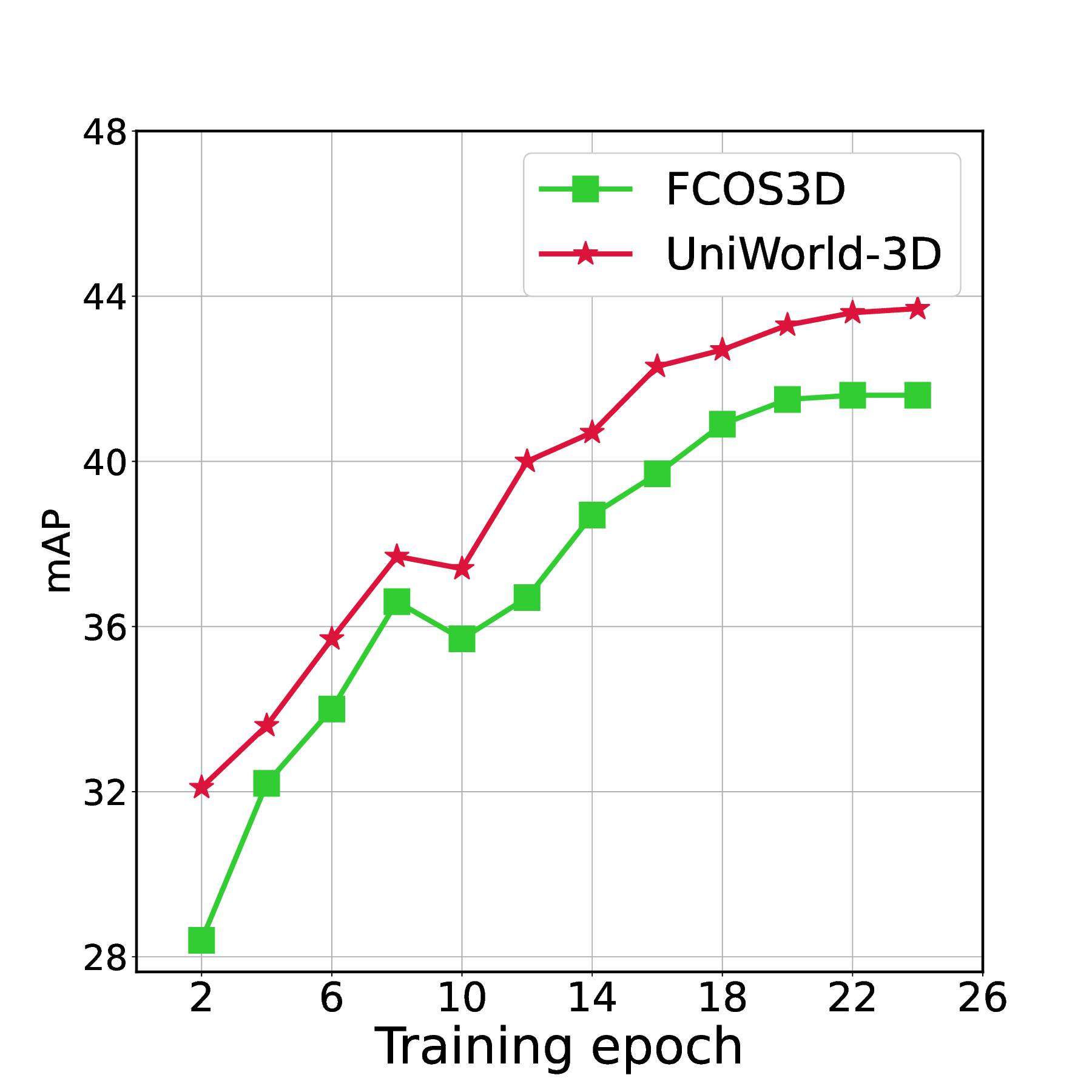}
		\captionof{figure}{Performance curves.
		}
		\label{fig:epoch}
	\end{minipage}%
	\begin{minipage}{.33\linewidth} 
		\includegraphics[width=.99\linewidth]{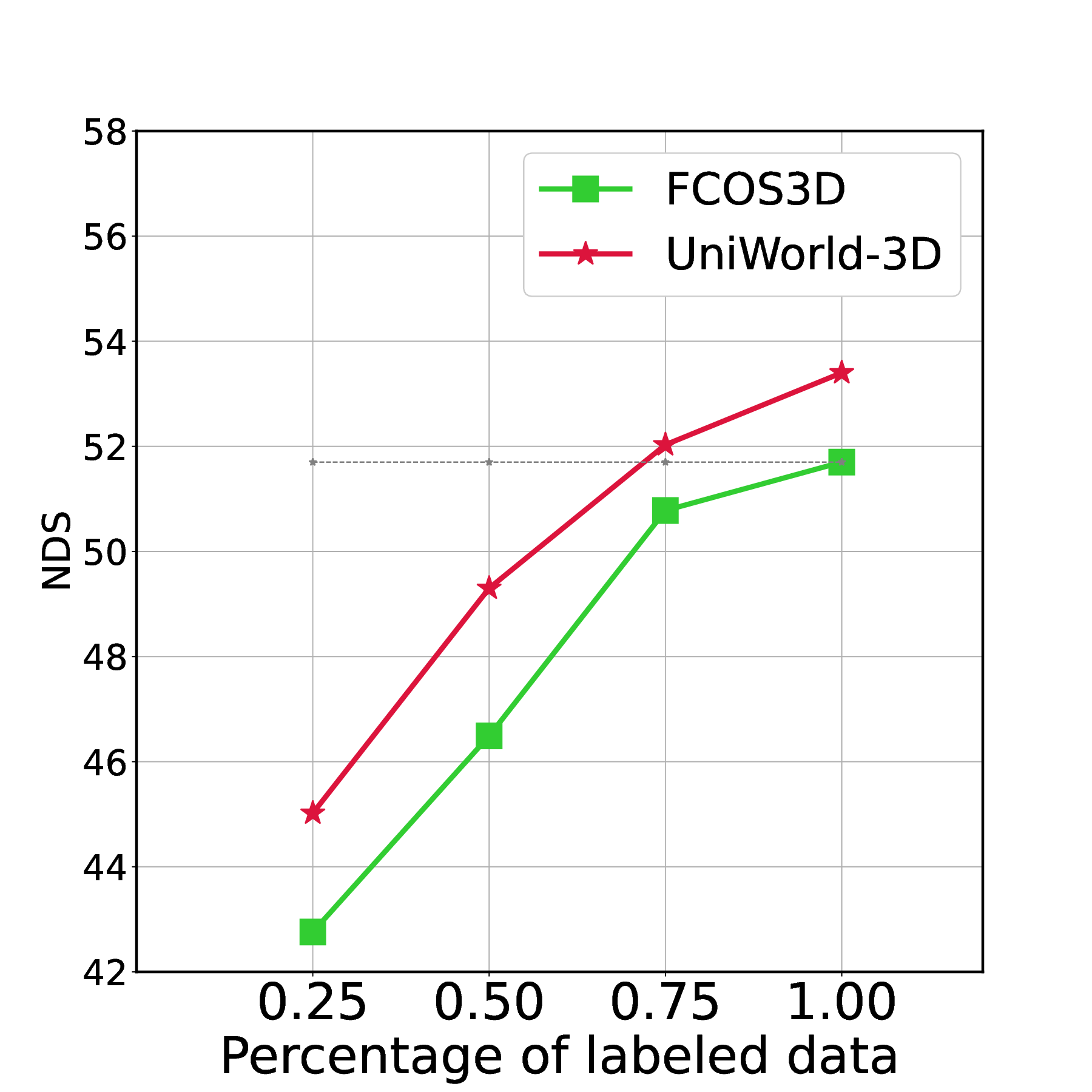}
		\captionof{figure}{Label-efficiency.
		}
		\label{fig:efficient}
	\end{minipage}%
	\begin{minipage}{.33\linewidth} 
		\resizebox{0.99\textwidth}{!}{
			\begin{tabular}{c|c|c|cc}
				\toprule
				\textbf{Frames}&\textbf{mAP}$\uparrow$&\textbf{NDS}$\uparrow$&\textbf{mATE}$\downarrow$\\
				\midrule
				0 &0.416&0.517&0.673 \\
				1 &0.417&0.523&0.661 \\
				3 &\textbf{0.438}&\textbf{0.534}&\textbf{0.656}\\ 
				5 &0.429&0.525&0.659  \\
				\bottomrule
			\end{tabular}
		}
		\captionof{table}{Impacts of multi-frame fusion. The occupancy labels were obtained by merging LiDAR point clouds from the current frame, preceding keyframes, succeeding keyframes, and their corresponding non-keyframes.}
		\label{tab:fusion}
	\end{minipage}%
	
	\begin{minipage}{.47\linewidth} 
		\centering
		\resizebox{0.85\textwidth}{!}{
			\begin{tabular}{c|c|c}
				\toprule
				\textbf{Methods} &\textbf{mAP}$\uparrow$&\textbf{NDS}$\uparrow$\\
				\midrule
				BEVFormer~\cite{bevformer}&0.352&0.423 \\
				BEVDistill~\cite{bevdistill}&0.386&0.457\\
				\midrule
				UniWorld-3D&\textbf{0.389}&\textbf{0.459}\\
				\bottomrule
		\end{tabular}}
		\captionof{table}{Comparison with BEVDistill~\cite{bevdistill}.}
		\label{tab:kd}
	\end{minipage}%
	\begin{minipage}{.52\linewidth} 
		\centering
		\resizebox{0.99\textwidth}{!}{
			\begin{tabular}{c|c|c|cc}
				\toprule
				\textbf{Methods}&\textbf{mAP}$\uparrow$&\textbf{NDS}$\uparrow$&\textbf{mATE}$\downarrow$\\
				\midrule
				BEVFormer~\cite{bevformer} &0.416&0.517&0.673 \\
				UniWorld-3D  &0.438&0.534&0.656  \\
				\midrule
				Supervision &\textbf{0.445}&\textbf{0.544}&\textbf{0.648} \\
				\bottomrule
		\end{tabular}}
		\captionof{table}{Semantic occupancy supervision.}
		\label{tab:supervision}
	\end{minipage}%
	
	\subsection{Ablation Studies}
	In this section, we perform thorough ablation experiments to investigate the individual components of our unified pre-training method UniWorld-3D with multi-camera 3D object detector BEVFormer~\cite{bevformer} on nuScenes \emph{val} set. 
	\subsubsection{Data-efficient Learner}
	
	Fine-tuning models using limited labeled data is made possible through pre-training. In order to evaluate the data efficiency of UniWorld-3D, we conducted experiments using varying amounts of labeled data for fine-tuning. BEVFormer~\cite{bevformer} was utilized as the backbone and assessed the detection performance of the model on the nuScenes validation set. The results, as depicted in Figure~\ref{fig:efficient}, demonstrate that when BEVFormer is trained with 75\% of the labeled data, it achieves the same performance as it trained on the complete dataset. Moreover, even with only 25\% of the samples available for fine-tuning, our UniWorld model outperforms BEVFormer by 1\% in mAP, highlighting its remarkable data efficiency and its potential to reduce the reliance on expensive human-annotated 3D data.
	
	\subsubsection{Multi-frame Fusion}
	
	
	We conducted an analysis of the influence of the number of fused LiDAR frames on the pre-trained model. As more frames were fused, the density of the point clouds increased. Our comparison included single-frame fusion, 3-frame fusion and 5-frame fusion (including the corresponding non-key frames) and the results are presented in Table~\ref{tab:fusion}. It is clear that the model's accuracy initially improved with an increasing number of fused point clouds but started to decline afterwards. This finding suggests that fusing multiple frames of point clouds can enhance the effectiveness of the pre-trained model. However, it is important to note that excessive fusion of frames introduces uncertainty due to the presence of dynamic objects. This uncertainty can lead to errors in the fusion process and subsequently lower the accuracy of the model.
	
	\subsubsection{Explicit Semantic Supervision}
	
	\begin{figure*}[t]
		\centering
		\includegraphics[width=0.98\textwidth]{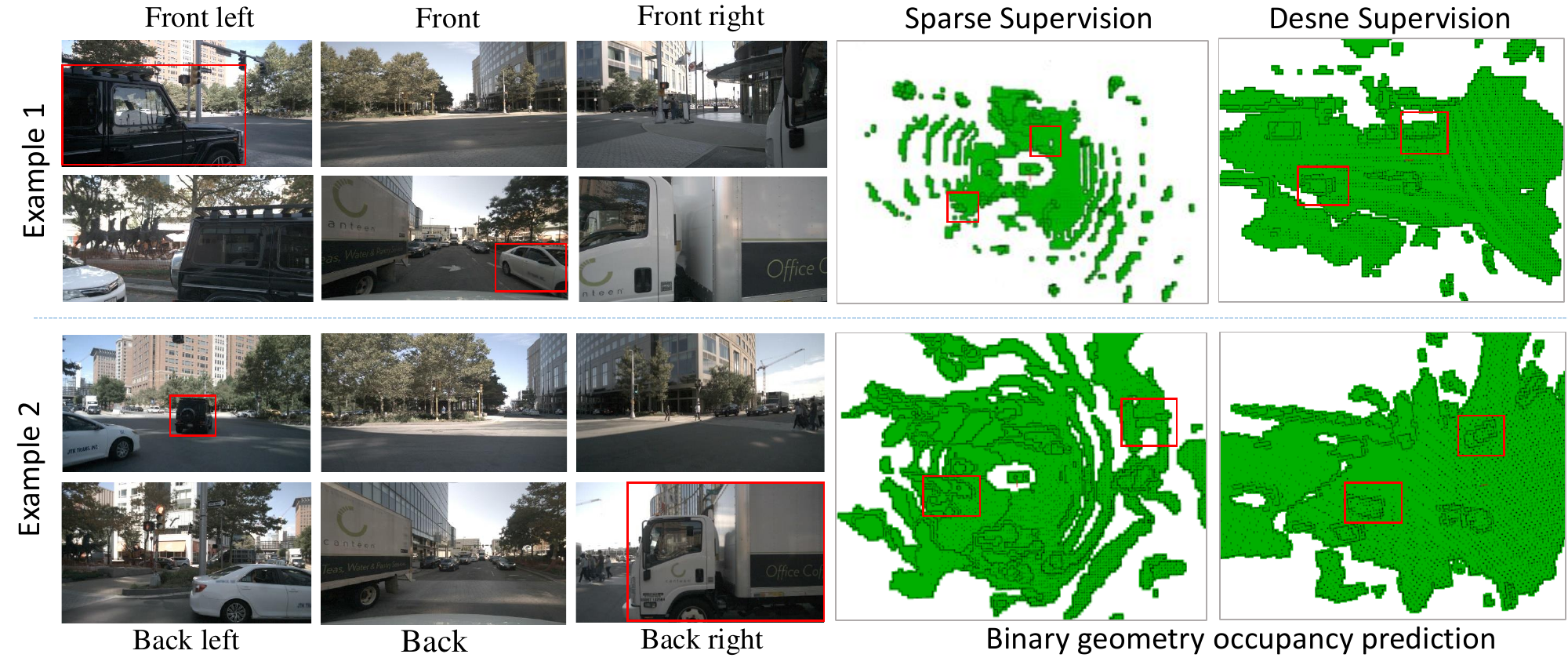} 
		\caption{Visualization of scene reconstruction via occupancy prediction.}
		\label{fig:show}
	\end{figure*}
	Labeled 3D data can be utilized to handle dynamic objects separately during the point cloud fusion process, resulting in more precise occupancy grid ground truth for multi-frame fusion. Subsequently, we examined the impact of explicit occupancy grid prediction on the model's performance. The results in Table~\ref{tab:supervision} demonstrate that incorporating explicit supervision leads to a notable improvement of 3\% in mAP and NDS compared to BEVFormer~\cite{bevformer}. Furthermore, when compared to unlabeled multi-frame fused point cloud pre-training, there is a 1\% increase in mAP. These findings highlight the potential of leveraging labeled data for explicit occupancy prediction supervision. Moreover, they further support the proposition that occupancy prediction enables the model to learn the data distribution of the entire 3D scene, thereby enhancing the accuracy of downstream tasks.
	
	%
	\subsection{Qualitative Evaluation}
	As shown in Figure~\ref{fig:show}, we present several reconstructed scenes. It can be observed that using single-frame point clouds as the supervision for occupancy grid generation results in incomplete reconstructions due to the sparse nature of the LiDAR point clouds. On the other hand, using three keyframes and their corresponding non-key frames as the supervision information allows for more complete reconstructions of scenes. 
	
	\textbf{Limitations.} Although our multi-camera unified pre-training approach has demonstrated promising results, there are several limitations to consider:
	(1) Currently, UniWorld-4D's performance in the 3D object detection task is slightly inferior to UniWorld-3D. It is necessary to address the issue of decreased 3D static detection performance caused by temporal information prediction.
	(2) The 3D convolutions in the decoder limit its applicability to tasks requiring high-resolution occupancy reconstruction. We will explore the cascade refine strategy. 
	(3) We rely on LiDAR to obtain ground truth occupancy grids. In the future, we will explore the NeRF~\cite{nerf,nerf2,block-nerf,dnerf} and MVS~\cite{mvs,mvsnet,aa-rmvsnet,bi} algorithms to reconstruct 3D scenes solely from multi-view images and obtain ground truth data.

	\section{Conclusion}
	We introduce a unified pre-training algorithm founded on 4D occupancy prediction to estimate missing information concerning the 3D world state and predict plausible future states of the 4D world. This approach has showcased remarkable efficacy across diverse autonomous driving tasks, such as motion prediction, multi-camera 3D object detection and surrounding semantic scene completion. Pre-training via World Model with unlabeled image-LiDAR pairs offers promising opportunities for reducing the 25\% dependency on annotated 3D data and establishing a Foundational Model for autonomous driving and robotics. Future work should focus on addressing the limitations mentioned and further improving the performance and applicability of our approach in real-world autonomous driving scenarios.

	%
	%
	%
	%
	
	\bibliography{example}  

\begin{thebibliography}{81}
\providecommand{\natexlab}[1]{#1}
\providecommand{\url}[1]{\texttt{#1}}
\expandafter\ifx\csname urlstyle\endcsname\relax
  \providecommand{\doi}[1]{doi: #1}\else
  \providecommand{\doi}{doi: \begingroup \urlstyle{rm}\Url}\fi

\bibitem[Elfes(1989)]{occupancy}
A.~Elfes.
\newblock Using occupancy grids for mobile robot perception and navigation.
\newblock \emph{Computer}, 22\penalty0 (6):\penalty0 46--57, 1989.

\bibitem[Ma et~al.(2022)Ma, Wang, Bai, Yang, Hou, Wang, Qiao, Yang, Manocha,
  and Zhu]{survey1}
Y.~Ma, T.~Wang, X.~Bai, H.~Yang, Y.~Hou, Y.~Wang, Y.~Qiao, R.~Yang, D.~Manocha,
  and X.~Zhu.
\newblock Vision-centric bev perception: A survey.
\newblock \emph{arXiv preprint arXiv:2208.02797}, 2022.

\bibitem[Li et~al.(2022)Li, Sima, Dai, Wang, Lu, Wang, Xie, Li, Deng, Tian,
  et~al.]{survey2}
H.~Li, C.~Sima, J.~Dai, W.~Wang, L.~Lu, H.~Wang, E.~Xie, Z.~Li, H.~Deng,
  H.~Tian, et~al.
\newblock Delving into the devils of bird's-eye-view perception: A review,
  evaluation and recipe.
\newblock \emph{arXiv preprint arXiv:2209.05324}, 2022.

\bibitem[Liang et~al.(2022)Liang, Xie, Yu, Xia, Lin, Wang, Tang, Wang, and
  Tang]{bevfusion}
T.~Liang, H.~Xie, K.~Yu, Z.~Xia, Z.~Lin, Y.~Wang, T.~Tang, B.~Wang, and
  Z.~Tang.
\newblock Bevfusion: A simple and robust lidar-camera fusion framework.
\newblock \emph{arXiv preprint arXiv:2205.13790}, 2022.

\bibitem[Chen et~al.(2023)Chen, Zhang, Wang, Wang, and Zhao]{chen2023futr3d}
X.~Chen, T.~Zhang, Y.~Wang, Y.~Wang, and H.~Zhao.
\newblock Futr3d: A unified sensor fusion framework for 3d detection.
\newblock In \emph{Proceedings of the IEEE/CVF Conference on Computer Vision
  and Pattern Recognition}, pages 172--181, 2023.

\bibitem[Li et~al.(2022)Li, Chen, Qi, Li, Sun, and Jia]{li2022unifying}
Y.~Li, Y.~Chen, X.~Qi, Z.~Li, J.~Sun, and J.~Jia.
\newblock Unifying voxel-based representation with transformer for 3d object
  detection.
\newblock \emph{arXiv preprint arXiv:2206.00630}, 2022.

\bibitem[Wang et~al.(2022)Wang, Guizilini, Zhang, Wang, Zhao, and
  Solomon]{detr3d}
Y.~Wang, V.~C. Guizilini, T.~Zhang, Y.~Wang, H.~Zhao, and J.~Solomon.
\newblock Detr3d: 3d object detection from multi-view images via 3d-to-2d
  queries.
\newblock In \emph{Conference on Robot Learning}, pages 180--191. PMLR, 2022.

\bibitem[Li et~al.(2022)Li, Wang, Li, Xie, Sima, Lu, Yu, and Dai]{bevformer}
Z.~Li, W.~Wang, H.~Li, E.~Xie, C.~Sima, T.~Lu, Q.~Yu, and J.~Dai.
\newblock Bevformer: Learning bird's-eye-view representation from multi-camera
  images via spatiotemporal transformers.
\newblock \emph{arXiv preprint arXiv:2203.17270}, 2022.

\bibitem[Huang et~al.(2021)Huang, Huang, Zhu, and Du]{bevdet}
J.~Huang, G.~Huang, Z.~Zhu, and D.~Du.
\newblock Bevdet: High-performance multi-camera 3d object detection in
  bird-eye-view.
\newblock \emph{arXiv preprint arXiv:2112.11790}, 2021.

\bibitem[Li et~al.(2022)Li, Ge, Yu, Yang, Wang, Shi, Sun, and Li]{bevdepth}
Y.~Li, Z.~Ge, G.~Yu, J.~Yang, Z.~Wang, Y.~Shi, J.~Sun, and Z.~Li.
\newblock Bevdepth: Acquisition of reliable depth for multi-view 3d object
  detection.
\newblock \emph{arXiv preprint arXiv:2206.10092}, 2022.

\bibitem[Xie et~al.(2022)Xie, Yu, Zhou, Philion, Anandkumar, Fidler, Luo, and
  Alvarez]{m2bev}
E.~Xie, Z.~Yu, D.~Zhou, J.~Philion, A.~Anandkumar, S.~Fidler, P.~Luo, and J.~M.
  Alvarez.
\newblock M\^{} 2bev: Multi-camera joint 3d detection and segmentation with
  unified birds-eye view representation.
\newblock \emph{arXiv preprint arXiv:2204.05088}, 2022.

\bibitem[Yang et~al.(2023)Yang, Chen, Tian, Tao, Zhu, Zhang, Huang, Li, Qiao,
  Lu, et~al.]{bevformerv2}
C.~Yang, Y.~Chen, H.~Tian, C.~Tao, X.~Zhu, Z.~Zhang, G.~Huang, H.~Li, Y.~Qiao,
  L.~Lu, et~al.
\newblock Bevformer v2: Adapting modern image backbones to bird's-eye-view
  recognition via perspective supervision.
\newblock In \emph{Proceedings of the IEEE/CVF Conference on Computer Vision
  and Pattern Recognition}, pages 17830--17839, 2023.

\bibitem[Deng et~al.(2009)Deng, Dong, Socher, Li, Li, and Fei-Fei]{imagenet}
J.~Deng, W.~Dong, R.~Socher, L.-J. Li, K.~Li, and L.~Fei-Fei.
\newblock Imagenet: A large-scale hierarchical image database.
\newblock In \emph{2009 IEEE conference on computer vision and pattern
  recognition}, pages 248--255. Ieee, 2009.

\bibitem[Park et~al.(2021)Park, Ambrus, Guizilini, Li, and Gaidon]{dd3d}
D.~Park, R.~Ambrus, V.~Guizilini, J.~Li, and A.~Gaidon.
\newblock Is pseudo-lidar needed for monocular 3d object detection?
\newblock In \emph{Proceedings of the IEEE/CVF International Conference on
  Computer Vision}, pages 3142--3152, 2021.

\bibitem[Shi et~al.(2023)Shi, Jiang, Li, Wen, Qian, Yang, Wang, and
  Yang]{occ_survey}
Y.~Shi, K.~Jiang, J.~Li, J.~Wen, Z.~Qian, M.~Yang, K.~Wang, and D.~Yang.
\newblock Grid-centric traffic scenario perception for autonomous driving: A
  comprehensive review.
\newblock \emph{arXiv preprint arXiv:2303.01212}, 2023.

\bibitem[Philion and Fidler(2020)]{lss}
J.~Philion and S.~Fidler.
\newblock Lift, splat, shoot: Encoding images from arbitrary camera rigs by
  implicitly unprojecting to 3d.
\newblock In \emph{European Conference on Computer Vision}, pages 194--210.
  Springer, 2020.

\bibitem[Caesar et~al.(2020)Caesar, Bankiti, Lang, Vora, Liong, Xu, Krishnan,
  Pan, Baldan, and Beijbom]{nuscenes}
H.~Caesar, V.~Bankiti, A.~H. Lang, S.~Vora, V.~E. Liong, Q.~Xu, A.~Krishnan,
  Y.~Pan, G.~Baldan, and O.~Beijbom.
\newblock nuscenes: A multimodal dataset for autonomous driving.
\newblock In \emph{Proceedings of the IEEE/CVF conference on computer vision
  and pattern recognition}, pages 11621--11631, 2020.

\bibitem[Zhang et~al.(2022)Zhang, Zhu, Zheng, Huang, Huang, Zhou, and
  Lu]{zhang2022beverse}
Y.~Zhang, Z.~Zhu, W.~Zheng, J.~Huang, G.~Huang, J.~Zhou, and J.~Lu.
\newblock Beverse: Unified perception and prediction in birds-eye-view for
  vision-centric autonomous driving.
\newblock \emph{arXiv preprint arXiv:2205.09743}, 2022.

\bibitem[Jiang et~al.(2022)Jiang, Zhang, Miao, Zhu, Gao, Hu, and
  Jiang]{jiang2022polarformer}
Y.~Jiang, L.~Zhang, Z.~Miao, X.~Zhu, J.~Gao, W.~Hu, and Y.-G. Jiang.
\newblock Polarformer: Multi-camera 3d object detection with polar
  transformers.
\newblock \emph{arXiv preprint arXiv:2206.15398}, 2022.

\bibitem[Hu et~al.(2022)Hu, Chen, Wu, Li, Yan, and Tao]{hu2022st}
S.~Hu, L.~Chen, P.~Wu, H.~Li, J.~Yan, and D.~Tao.
\newblock St-p3: End-to-end vision-based autonomous driving via
  spatial-temporal feature learning.
\newblock In \emph{Computer Vision--ECCV 2022: 17th European Conference, Tel
  Aviv, Israel, October 23--27, 2022, Proceedings, Part XXXVIII}, pages
  533--549. Springer, 2022.

\bibitem[Xie et~al.(2023)Xie, Kong, Zhang, Ren, Pan, Chen, and
  Liu]{xie2023robobev}
S.~Xie, L.~Kong, W.~Zhang, J.~Ren, L.~Pan, K.~Chen, and Z.~Liu.
\newblock Robobev: Towards robust bird's eye view perception under corruptions.
\newblock \emph{arXiv preprint arXiv:2304.06719}, 2023.

\bibitem[Chen et~al.(2022)Chen, Li, Zhang, Fang, Jiang, and
  Zhao]{chen2022graph}
Z.~Chen, Z.~Li, S.~Zhang, L.~Fang, Q.~Jiang, and F.~Zhao.
\newblock Graph-detr3d: rethinking overlapping regions for multi-view 3d object
  detection.
\newblock In \emph{Proceedings of the 30th ACM International Conference on
  Multimedia}, pages 5999--6008, 2022.

\bibitem[Li et~al.(2023)Li, Huang, Chen, Cui, Liang, Shen, Liu, Xie, Sheng,
  Ouyang, et~al.]{fastbev}
Y.~Li, B.~Huang, Z.~Chen, Y.~Cui, F.~Liang, M.~Shen, F.~Liu, E.~Xie, L.~Sheng,
  W.~Ouyang, et~al.
\newblock Fast-bev: A fast and strong bird's-eye view perception baseline.
\newblock \emph{arXiv preprint arXiv:2301.12511}, 2023.

\bibitem[Li et~al.(2022)Li, Wang, Wang, and Zhao]{li2022hdmapnet}
Q.~Li, Y.~Wang, Y.~Wang, and H.~Zhao.
\newblock Hdmapnet: An online hd map construction and evaluation framework.
\newblock In \emph{2022 International Conference on Robotics and Automation
  (ICRA)}, pages 4628--4634. IEEE, 2022.

\bibitem[Liao et~al.(2022)Liao, Chen, Wang, Cheng, Zhang, Liu, and
  Huang]{liao2022maptr}
B.~Liao, S.~Chen, X.~Wang, T.~Cheng, Q.~Zhang, W.~Liu, and C.~Huang.
\newblock Maptr: Structured modeling and learning for online vectorized hd map
  construction.
\newblock \emph{arXiv preprint arXiv:2208.14437}, 2022.

\bibitem[Doll et~al.(2022)Doll, Schulz, Schneider, Benzin, Enzweiler, and
  Lensch]{doll2022spatialdetr}
S.~Doll, R.~Schulz, L.~Schneider, V.~Benzin, M.~Enzweiler, and H.~P. Lensch.
\newblock Spatialdetr: Robust scalable transformer-based 3d object detection
  from multi-view camera images with global cross-sensor attention.
\newblock In \emph{Computer Vision--ECCV 2022: 17th European Conference, Tel
  Aviv, Israel, October 23--27, 2022, Proceedings, Part XXXIX}, pages 230--245.
  Springer, 2022.

\bibitem[Zhou et~al.(2022)Zhou, Ge, Li, and Zhang]{zhou2022matrixvt}
H.~Zhou, Z.~Ge, Z.~Li, and X.~Zhang.
\newblock Matrixvt: Efficient multi-camera to bev transformation for 3d
  perception.
\newblock \emph{arXiv preprint arXiv:2211.10593}, 2022.

\bibitem[Li et~al.(2022)Li, Bao, Ge, Yang, Sun, and Li]{bevstereo}
Y.~Li, H.~Bao, Z.~Ge, J.~Yang, J.~Sun, and Z.~Li.
\newblock Bevstereo: Enhancing depth estimation in multi-view 3d object
  detection with dynamic temporal stereo.
\newblock \emph{arXiv preprint arXiv:2209.10248}, 2022.

\bibitem[Wang et~al.(2022)Wang, Min, Ge, Li, Li, Yang, and Huang]{sts}
Z.~Wang, C.~Min, Z.~Ge, Y.~Li, Z.~Li, H.~Yang, and D.~Huang.
\newblock Sts: Surround-view temporal stereo for multi-view 3d detection.
\newblock \emph{arXiv preprint arXiv:2208.10145}, 2022.

\bibitem[Park et~al.(2022)Park, Xu, Yang, Keutzer, Kitani, Tomizuka, and
  Zhan]{solofusion}
J.~Park, C.~Xu, S.~Yang, K.~Keutzer, K.~Kitani, M.~Tomizuka, and W.~Zhan.
\newblock Time will tell: New outlooks and a baseline for temporal multi-view
  3d object detection.
\newblock \emph{arXiv preprint arXiv:2210.02443}, 2022.

\bibitem[Han et~al.(2023)Han, Sun, Ge, Yang, Dong, Zhou, Mao, Peng, and
  Zhang]{videobev}
C.~Han, J.~Sun, Z.~Ge, J.~Yang, R.~Dong, H.~Zhou, W.~Mao, Y.~Peng, and
  X.~Zhang.
\newblock Exploring recurrent long-term temporal fusion for multi-view 3d
  perception.
\newblock \emph{arXiv preprint arXiv:2303.05970}, 2023.

\bibitem[Liu et~al.(2022)Liu, Wang, Zhang, and Sun]{petr}
Y.~Liu, T.~Wang, X.~Zhang, and J.~Sun.
\newblock Petr: Position embedding transformation for multi-view 3d object
  detection.
\newblock \emph{arXiv preprint arXiv:2203.05625}, 2022.

\bibitem[Hu et~al.(2023)Hu, Yang, Chen, Li, Sima, Zhu, Chai, Du, Lin, Wang, Lu,
  Jia, Liu, Dai, Qiao, and Li]{uniad}
Y.~Hu, J.~Yang, L.~Chen, K.~Li, C.~Sima, X.~Zhu, S.~Chai, S.~Du, T.~Lin,
  W.~Wang, L.~Lu, X.~Jia, Q.~Liu, J.~Dai, Y.~Qiao, and H.~Li.
\newblock Planning-oriented autonomous driving.
\newblock In \emph{Proceedings of the IEEE/CVF Conference on Computer Vision
  and Pattern Recognition}, 2023.

\bibitem[Doersch et~al.(2015)Doersch, Gupta, and Efros]{patch_id}
C.~Doersch, A.~Gupta, and A.~A. Efros.
\newblock Unsupervised visual representation learning by context prediction.
\newblock In \emph{Proceedings of the IEEE international conference on computer
  vision}, pages 1422--1430, 2015.

\bibitem[Carlucci et~al.(2019)Carlucci, D'Innocente, Bucci, Caputo, and
  Tommasi]{jigsaw}
F.~M. Carlucci, A.~D'Innocente, S.~Bucci, B.~Caputo, and T.~Tommasi.
\newblock Domain generalization by solving jigsaw puzzles.
\newblock In \emph{Proceedings of the IEEE/CVF Conference on Computer Vision
  and Pattern Recognition}, pages 2229--2238, 2019.

\bibitem[Caron et~al.(2018)Caron, Bojanowski, Joulin, and Douze]{deepcluster}
M.~Caron, P.~Bojanowski, A.~Joulin, and M.~Douze.
\newblock Deep clustering for unsupervised learning of visual features.
\newblock In \emph{Proceedings of the European conference on computer vision
  (ECCV)}, pages 132--149, 2018.

\bibitem[Caron et~al.(2020)Caron, Misra, Mairal, Goyal, Bojanowski, and
  Joulin]{swav}
M.~Caron, I.~Misra, J.~Mairal, P.~Goyal, P.~Bojanowski, and A.~Joulin.
\newblock Unsupervised learning of visual features by contrasting cluster
  assignments.
\newblock \emph{Advances in Neural Information Processing Systems},
  33:\penalty0 9912--9924, 2020.

\bibitem[He et~al.(2020)He, Fan, Wu, Xie, and Girshick]{moco}
K.~He, H.~Fan, Y.~Wu, S.~Xie, and R.~Girshick.
\newblock Momentum contrast for unsupervised visual representation learning.
\newblock In \emph{Proceedings of the IEEE/CVF conference on computer vision
  and pattern recognition}, pages 9729--9738, 2020.

\bibitem[Grill et~al.(2020)Grill, Strub, Altch{\'e}, Tallec, Richemond,
  Buchatskaya, Doersch, Avila~Pires, Guo, Gheshlaghi~Azar, et~al.]{byol}
J.-B. Grill, F.~Strub, F.~Altch{\'e}, C.~Tallec, P.~Richemond, E.~Buchatskaya,
  C.~Doersch, B.~Avila~Pires, Z.~Guo, M.~Gheshlaghi~Azar, et~al.
\newblock Bootstrap your own latent-a new approach to self-supervised learning.
\newblock \emph{Advances in neural information processing systems},
  33:\penalty0 21271--21284, 2020.

\bibitem[He et~al.(2022)He, Chen, Xie, Li, Doll{\'a}r, and Girshick]{mae}
K.~He, X.~Chen, S.~Xie, Y.~Li, P.~Doll{\'a}r, and R.~Girshick.
\newblock Masked autoencoders are scalable vision learners.
\newblock In \emph{Proceedings of the IEEE/CVF Conference on Computer Vision
  and Pattern Recognition}, pages 16000--16009, 2022.

\bibitem[Bao et~al.(2021)Bao, Dong, Piao, and Wei]{beit}
H.~Bao, L.~Dong, S.~Piao, and F.~Wei.
\newblock Beit: Bert pre-training of image transformers.
\newblock \emph{arXiv preprint arXiv:2106.08254}, 2021.

\bibitem[Min et~al.(2022)Min, Zhao, Xiao, Nie, and Dai]{voxel-mae}
C.~Min, D.~Zhao, L.~Xiao, Y.~Nie, and B.~Dai.
\newblock Voxel-mae: Masked autoencoders for pre-training large-scale point
  clouds.
\newblock \emph{arXiv preprint arXiv:2206.09900}, 2022.

\bibitem[Boulch et~al.(2022)Boulch, Sautier, Michele, Puy, and Marlet]{also}
A.~Boulch, C.~Sautier, B.~Michele, G.~Puy, and R.~Marlet.
\newblock Also: Automotive lidar self-supervision by occupancy estimation.
\newblock \emph{arXiv preprint arXiv:2212.05867}, 2022.

\bibitem[Min et~al.(2023)Min, Xu, Zhao, Xiao, Nie, and Dai]{occbev}
C.~Min, X.~Xu, D.~Zhao, L.~Xiao, Y.~Nie, and B.~Dai.
\newblock Occ-bev: Multi-camera unified pre-training via 3d scene
  reconstruction.
\newblock \emph{arXiv preprint arXiv:2305.18829}, 2023.

\bibitem[Achinstein(1983)]{achinstein1983nature}
P.~Achinstein.
\newblock \emph{The nature of explanation}.
\newblock Oxford University Press, USA, 1983.

\bibitem[Ha and Schmidhuber(2018{\natexlab{a}})]{world_models}
D.~Ha and J.~Schmidhuber.
\newblock World models.
\newblock \emph{arXiv preprint arXiv:1803.10122}, 2018{\natexlab{a}}.

\bibitem[Ha and Schmidhuber(2018{\natexlab{b}})]{recurrent_wm}
D.~Ha and J.~Schmidhuber.
\newblock Recurrent world models facilitate policy evolution.
\newblock \emph{Advances in neural information processing systems}, 31,
  2018{\natexlab{b}}.

\bibitem[Hafner et~al.(2020)Hafner, Lillicrap, Norouzi, and Ba]{mastering_wm}
D.~Hafner, T.~Lillicrap, M.~Norouzi, and J.~Ba.
\newblock Mastering atari with discrete world models.
\newblock \emph{arXiv preprint arXiv:2010.02193}, 2020.

\bibitem[Schrittwieser et~al.(2020)Schrittwieser, Antonoglou, Hubert, Simonyan,
  Sifre, Schmitt, Guez, Lockhart, Hassabis, Graepel, et~al.]{mastering_wm2}
J.~Schrittwieser, I.~Antonoglou, T.~Hubert, K.~Simonyan, L.~Sifre, S.~Schmitt,
  A.~Guez, E.~Lockhart, D.~Hassabis, T.~Graepel, et~al.
\newblock Mastering atari, go, chess and shogi by planning with a learned
  model.
\newblock \emph{Nature}, 588\penalty0 (7839):\penalty0 604--609, 2020.

\bibitem[Babaeizadeh et~al.(2017)Babaeizadeh, Finn, Erhan, Campbell, and
  Levine]{babaeizadeh2017stochastic}
M.~Babaeizadeh, C.~Finn, D.~Erhan, R.~H. Campbell, and S.~Levine.
\newblock Stochastic variational video prediction.
\newblock \emph{arXiv preprint arXiv:1710.11252}, 2017.

\bibitem[Denton and Fergus(2018)]{denton2018stochastic}
E.~Denton and R.~Fergus.
\newblock Stochastic video generation with a learned prior.
\newblock In \emph{International conference on machine learning}, pages
  1174--1183. PMLR, 2018.

\bibitem[Franceschi et~al.(2020)Franceschi, Delasalles, Chen, Lamprier, and
  Gallinari]{franceschi2020stochastic}
J.-Y. Franceschi, E.~Delasalles, M.~Chen, S.~Lamprier, and P.~Gallinari.
\newblock Stochastic latent residual video prediction.
\newblock In \emph{International Conference on Machine Learning}, pages
  3233--3246. PMLR, 2020.

\bibitem[Hu et~al.(2022)Hu, Corrado, Griffiths, Murez, Gurau, Yeo, Kendall,
  Cipolla, and Shotton]{hu2022model}
A.~Hu, G.~Corrado, N.~Griffiths, Z.~Murez, C.~Gurau, H.~Yeo, A.~Kendall,
  R.~Cipolla, and J.~Shotton.
\newblock Model-based imitation learning for urban driving.
\newblock \emph{Advances in Neural Information Processing Systems},
  35:\penalty0 20703--20716, 2022.

\bibitem[He et~al.(2016)He, Zhang, Ren, and Sun]{resnet}
K.~He, X.~Zhang, S.~Ren, and J.~Sun.
\newblock Deep residual learning for image recognition.
\newblock In \emph{Proceedings of the IEEE conference on computer vision and
  pattern recognition}, pages 770--778, 2016.

\bibitem[Yan et~al.(2018)Yan, Mao, and Li]{second}
Y.~Yan, Y.~Mao, and B.~Li.
\newblock Second: Sparsely embedded convolutional detection.
\newblock \emph{Sensors}, 18\penalty0 (10):\penalty0 3337, 2018.

\bibitem[Shi et~al.(2020)Shi, Guo, Jiang, Wang, Shi, Wang, and Li]{pv_rcnn}
S.~Shi, C.~Guo, L.~Jiang, Z.~Wang, J.~Shi, X.~Wang, and H.~Li.
\newblock Pv-rcnn: Point-voxel feature set abstraction for 3d object detection.
\newblock In \emph{Proceedings of the IEEE/CVF Conference on Computer Vision
  and Pattern Recognition}, pages 10529--10538, 2020.

\bibitem[Yin et~al.(2021)Yin, Zhou, and Krahenbuhl]{centerpoint}
T.~Yin, X.~Zhou, and P.~Krahenbuhl.
\newblock Center-based 3d object detection and tracking.
\newblock In \emph{Proceedings of the IEEE/CVF conference on computer vision
  and pattern recognition}, pages 11784--11793, 2021.

\bibitem[Zhu et~al.(2021)Zhu, Zhou, Wang, Hong, Ma, Li, Li, and
  Lin]{cylinder3d}
X.~Zhu, H.~Zhou, T.~Wang, F.~Hong, Y.~Ma, W.~Li, H.~Li, and D.~Lin.
\newblock Cylindrical and asymmetrical 3d convolution networks for lidar
  segmentation.
\newblock In \emph{Proceedings of the IEEE/CVF conference on computer vision
  and pattern recognition}, pages 9939--9948, 2021.

\bibitem[Mildenhall et~al.(2021)Mildenhall, Srinivasan, Tancik, Barron,
  Ramamoorthi, and Ng]{nerf}
B.~Mildenhall, P.~P. Srinivasan, M.~Tancik, J.~T. Barron, R.~Ramamoorthi, and
  R.~Ng.
\newblock Nerf: Representing scenes as neural radiance fields for view
  synthesis.
\newblock \emph{Communications of the ACM}, 65\penalty0 (1):\penalty0 99--106,
  2021.

\bibitem[Barron et~al.(2021)Barron, Mildenhall, Tancik, Hedman, Martin-Brualla,
  and Srinivasan]{nerf2}
J.~T. Barron, B.~Mildenhall, M.~Tancik, P.~Hedman, R.~Martin-Brualla, and P.~P.
  Srinivasan.
\newblock Mip-nerf: A multiscale representation for anti-aliasing neural
  radiance fields.
\newblock In \emph{Proceedings of the IEEE/CVF International Conference on
  Computer Vision}, pages 5855--5864, 2021.

\bibitem[Tancik et~al.(2022)Tancik, Casser, Yan, Pradhan, Mildenhall,
  Srinivasan, Barron, and Kretzschmar]{block-nerf}
M.~Tancik, V.~Casser, X.~Yan, S.~Pradhan, B.~Mildenhall, P.~P. Srinivasan,
  J.~T. Barron, and H.~Kretzschmar.
\newblock Block-nerf: Scalable large scene neural view synthesis.
\newblock In \emph{Proceedings of the IEEE/CVF Conference on Computer Vision
  and Pattern Recognition}, pages 8248--8258, 2022.

\bibitem[Pumarola et~al.(2021)Pumarola, Corona, Pons-Moll, and
  Moreno-Noguer]{dnerf}
A.~Pumarola, E.~Corona, G.~Pons-Moll, and F.~Moreno-Noguer.
\newblock D-nerf: Neural radiance fields for dynamic scenes.
\newblock In \emph{Proceedings of the IEEE/CVF Conference on Computer Vision
  and Pattern Recognition}, pages 10318--10327, 2021.

\bibitem[Zhu et~al.(2021)Zhu, Min, Wei, Chen, and Wang]{mvs}
Q.~Zhu, C.~Min, Z.~Wei, Y.~Chen, and G.~Wang.
\newblock Deep learning for multi-view stereo via plane sweep: A survey.
\newblock \emph{arXiv preprint arXiv:2106.15328}, 2021.

\bibitem[Yao et~al.(2018)Yao, Luo, Li, Fang, and Quan]{mvsnet}
Y.~Yao, Z.~Luo, S.~Li, T.~Fang, and L.~Quan.
\newblock Mvsnet: Depth inference for unstructured multi-view stereo.
\newblock In \emph{Proceedings of the European conference on computer vision
  (ECCV)}, pages 767--783, 2018.

\bibitem[Wei et~al.(2021)Wei, Zhu, Min, Chen, and Wang]{aa-rmvsnet}
Z.~Wei, Q.~Zhu, C.~Min, Y.~Chen, and G.~Wang.
\newblock Aa-rmvsnet: Adaptive aggregation recurrent multi-view stereo network.
\newblock In \emph{Proceedings of the IEEE/CVF International Conference on
  Computer Vision}, pages 6187--6196, 2021.

\bibitem[Wei et~al.(2022)Wei, Zhu, Min, Chen, and Wang]{bi}
Z.~Wei, Q.~Zhu, C.~Min, Y.~Chen, and G.~Wang.
\newblock Bidirectional hybrid lstm based recurrent neural network for
  multi-view stereo.
\newblock \emph{IEEE Transactions on Visualization and Computer Graphics},
  2022.

\bibitem[Huang et~al.(2023)Huang, Zheng, Zhang, Zhou, and Lu]{tpvformer}
Y.~Huang, W.~Zheng, Y.~Zhang, J.~Zhou, and J.~Lu.
\newblock Tri-perspective view for vision-based 3d semantic occupancy
  prediction.
\newblock \emph{arXiv preprint arXiv:2302.07817}, 2023.

\bibitem[Wang et~al.(2023)Wang, Zhu, Xu, Zhang, Wei, Chi, Ye, Du, Lu, and
  Wang]{openoccupancy}
X.~Wang, Z.~Zhu, W.~Xu, Y.~Zhang, Y.~Wei, X.~Chi, Y.~Ye, D.~Du, J.~Lu, and
  X.~Wang.
\newblock Openoccupancy: A large scale benchmark for surrounding semantic
  occupancy perception.
\newblock \emph{arXiv preprint arXiv:2303.03991}, 2023.

\bibitem[Tian et~al.(2023)Tian, Jiang, Yun, Wang, Wang, and Zhao]{occ3d}
X.~Tian, T.~Jiang, L.~Yun, Y.~Wang, Y.~Wang, and H.~Zhao.
\newblock Occ3d: A large-scale 3d occupancy prediction benchmark for autonomous
  driving.
\newblock \emph{arXiv preprint arXiv:2304.14365}, 2023.

\bibitem[Li et~al.(2023)Li, Yu, Choy, Xiao, Alvarez, Fidler, Feng, and
  Anandkumar]{li2023voxformer}
Y.~Li, Z.~Yu, C.~Choy, C.~Xiao, J.~M. Alvarez, S.~Fidler, C.~Feng, and
  A.~Anandkumar.
\newblock Voxformer: Sparse voxel transformer for camera-based 3d semantic
  scene completion.
\newblock In \emph{Proceedings of the IEEE/CVF Conference on Computer Vision
  and Pattern Recognition}, pages 9087--9098, 2023.

\bibitem[Wei et~al.(2023)Wei, Zhao, Zheng, Zhu, Zhou, and
  Lu]{wei2023surroundocc}
Y.~Wei, L.~Zhao, W.~Zheng, Z.~Zhu, J.~Zhou, and J.~Lu.
\newblock Surroundocc: Multi-camera 3d occupancy prediction for autonomous
  driving.
\newblock \emph{arXiv preprint arXiv:2303.09551}, 2023.

\bibitem[Zhang et~al.(2023)Zhang, Zhu, and Du]{zhang2023occformer}
Y.~Zhang, Z.~Zhu, and D.~Du.
\newblock Occformer: Dual-path transformer for vision-based 3d semantic
  occupancy prediction.
\newblock \emph{arXiv preprint arXiv:2304.05316}, 2023.

\bibitem[Miao et~al.(2023)Miao, Liu, Chen, Gong, Xu, Hu, and
  Zhou]{miao2023occdepth}
R.~Miao, W.~Liu, M.~Chen, Z.~Gong, W.~Xu, C.~Hu, and S.~Zhou.
\newblock Occdepth: A depth-aware method for 3d semantic scene completion.
\newblock \emph{arXiv preprint arXiv:2302.13540}, 2023.

\bibitem[Gan et~al.(2023)Gan, Mo, Xu, and Yokoya]{gan2023simple}
W.~Gan, N.~Mo, H.~Xu, and N.~Yokoya.
\newblock A simple attempt for 3d occupancy estimation in autonomous driving.
\newblock \emph{arXiv preprint arXiv:2303.10076}, 2023.

\bibitem[Chen et~al.(2022)Chen, Li, Zhang, Fang, Jiang, and Zhao]{bevdistill}
Z.~Chen, Z.~Li, S.~Zhang, L.~Fang, Q.~Jiang, and F.~Zhao.
\newblock Bevdistill: Cross-modal bev distillation for multi-view 3d object
  detection.
\newblock \emph{arXiv preprint arXiv:2211.09386}, 2022.

\bibitem[Huang et~al.(2022)Huang, Liu, Zhang, Zhang, Xu, Wang, and Liu]{tigbev}
P.~Huang, L.~Liu, R.~Zhang, S.~Zhang, X.~Xu, B.~Wang, and G.~Liu.
\newblock Tig-bev: Multi-view bev 3d object detection via target inner-geometry
  learning.
\newblock \emph{arXiv preprint arXiv:2212.13979}, 2022.

\bibitem[Liu et~al.(2023)Liu, Wang, Liu, Zhang, Liu, and Li]{geomim}
J.~Liu, T.~Wang, B.~Liu, Q.~Zhang, Y.~Liu, and H.~Li.
\newblock Towards better 3d knowledge transfer via masked image modeling for
  multi-view 3d understanding.
\newblock \emph{arXiv preprint arXiv:2303.11325}, 2023.

\bibitem[Zhang et~al.(2022)Zhang, Zhu, Zheng, Huang, Huang, Zhou, and
  Lu]{beverse}
Y.~Zhang, Z.~Zhu, W.~Zheng, J.~Huang, G.~Huang, J.~Zhou, and J.~Lu.
\newblock Beverse: Unified perception and prediction in birds-eye-view for
  vision-centric autonomous driving.
\newblock \emph{arXiv preprint arXiv:2205.09743}, 2022.

\bibitem[Wang et~al.(2021)Wang, Zhu, Pang, and Lin]{fcos3d}
T.~Wang, X.~Zhu, J.~Pang, and D.~Lin.
\newblock {FCOS3D: Fully} convolutional one-stage monocular 3d object
  detection.
\newblock In \emph{Proceedings of the IEEE/CVF International Conference on
  Computer Vision (ICCV) Workshops}, 2021.

\bibitem[occ(2023)]{occ}
Cvpr 2023 occupancy prediction challenge, 2023.
\newblock URL
  \url{https://github.com/CVPR2023-3D-Occupancy-Prediction/CVPR2023-3D-Occupancy-Prediction}.

\bibitem[Liu et~al.(2021)Liu, Lin, Cao, Hu, Wei, Zhang, Lin, and Guo]{swin}
Z.~Liu, Y.~Lin, Y.~Cao, H.~Hu, Y.~Wei, Z.~Zhang, S.~Lin, and B.~Guo.
\newblock Swin transformer: Hierarchical vision transformer using shifted
  windows.
\newblock In \emph{Proceedings of the IEEE/CVF international conference on
  computer vision}, pages 10012--10022, 2021.

\end{thebibliography}
	
\end{document}